# Line detection via a lightweight CNN with a Hough Layer

Lev Teplyakov[1,*], Kirill Kaymakov[1], Evgeny Shvets[1], Dmitry Nikolaev[1]
[1]Institute for Information Transmission Problems RAS, Moscow, Russia
[*]Corresponding author, teplyakov@iitp.ru

## ABSTRACT

Line detection is an important computer vision task traditionally solved by Hough Transform. With the advance of deep learning, however, trainable approaches to line detection became popular. In this paper we propose a lightweight CNN for line detection with an embedded parameter-free Hough layer, which allows the network neurons to have global strip-like receptive fields. We argue that traditional convolutional networks have two inherent problems when applied to the task of line detection and show how insertion of a Hough layer into the network solves them. Additionally, we point out some major inconsistencies in the current datasets used for line detection.

## 1. INTRODUCTION

Line detection (LD) is a well-studied, classic computer vision problem. LD algorithms are crucial for many computer vision tasks. Line detection acts as an intermediate step in some line segment detection (LSD) algorithms [1]; it is also used by itself to detect vanishing points [2] and road lanes [3,4] by autonomous ground vehicles; to detect power lines for unmanned aerial vehicles [5]; to estimate and rectify homography [6, 7] or skew [8] in document recognition.

The difference between the problems of line and line segment detection is that LD algorithms aim to find not individual continuous line segments, but rather lines that are "salient" as a whole, e.g. contain several line segments or points. For instance, a dashed line should trigger several line segment detections, one per dash; but only one line detection. Conversely, a dotted straight line could trigger a line detection, whereas it must not trigger a line segment detection.

Traditionally, LD has been solved by the usage of Hough Transform (HT) and its variations [9, 10]. Line detection via HT consists of three general steps: (i) extracting the feature image from the original image (e.g. by Canny edge detector [11]), (ii) mapping the feature image into a discrete parameter space (by the HT itself), (iii) finding peaks in the parameter space. Over the years, dozens of HT modifications have been proposed. A review on Hough Transform and its variations can be found in [12].

There are several problems with finding lines using Standard Hough Transform (SHT). The first problem is the presence of noise or textures in the image. See Fig.1 for example: the straight lines in the feature image cannot be distinguished due to the excessive amount of textures; this can create many false positive detections. In [13] authors consider this problem and combat it by downweighting the votes of features that are surrounded by textures (have strong surround suppression).

Let us consider the second problem of SHT for line detection. SHT converts a salient line in an image into a butterfly-like pattern (a cluster of cells with high vote count) in the Hough space, and deriving a single answer from this pattern is not straightforward. The exact shape of Radon (and Hough) transform peaks has been studied, e.g. in [14]; and it was shown that the peak does not always have a pronounced single pixel with maximal value. The issue is additionally aggravated when the image contains textures, noise or geometric distortions.

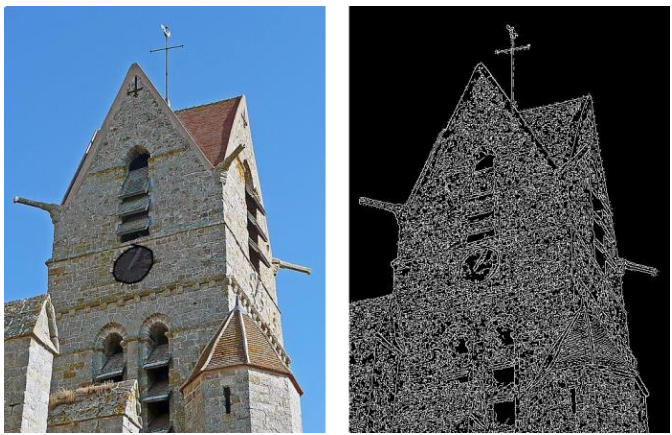 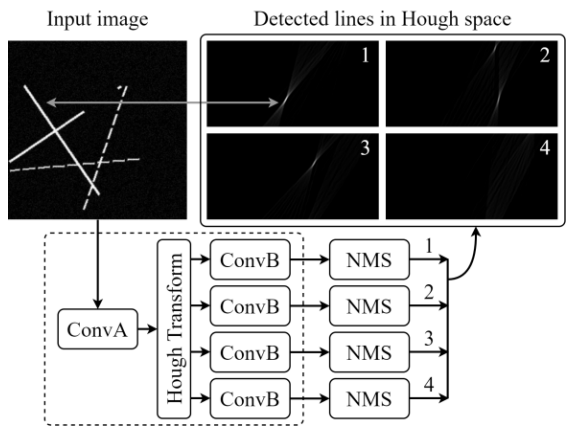

Figure 1. An image and the corresponding output of Canny edge detector.

Figure 2. General pipeline of the proposed approach for line detection, LNet.

Finally, there is the complexity problem. The naive HT (brute force summation over all lines) has $O(N^3)$ complexity (where N is the linear size of the image). Randomized HT and multiple other probabilistic approaches aim to improve the running time of the algorithm by processing only a subset of point (for an excellent review, see [12]). Another approach – one that relies on an alternative approximation of a line in the discrete image, but calculates the exact sums over them – is called Fast Hough Transform [15]. It speeds up the algorithm by (i) approximating the lines with dyadic patters, (ii) calculating the sums of pixels over common sub-patterns and (iii) reusing these intermediate results.

In this paper we propose a variation of HT–based line detection algorithm. Specifically, we propose a lightweight artificial neural network for line detection that has a few convolutional layers and a Fast Hough Transform layer and can be trained in an end-to-end manner.

We additionally argue that while purely convolutional networks have been very popular in the recent years and achieved state-of-the-art performance in many problems (including LSD - [16]), they have two inherent major drawbacks that make them inefficient for line detection. We show how inclusion of the differentiable Hough layer allows to solve these problems.

**Related research**

To our best knowledge, there are three papers that have proposed to embed a Hough layer into a neural network. The original idea belongs to [17] (with further research [18] from the same authors), where an end-to-end network with two Hough layers was used to detect vanishing points. In a very recent paper [19], Hough layer was inserted into an end-to-end network specifically for line detection; our work differs from it in two major points. Firstly, authors argue that they train a network to find "artistically important" lines in the image - with the purpose of automatic image enhancement. Therefore, the dataset and trained network cannot be used as a ready-to-use substitution for the Hough transform: the network is trained to find very specific lines in the image and is not a general-purpose line detector. Secondly, our network is hundreds time smaller (in both trainable parameters and computations), with only 4 layers with maximum of 16 channels and basically could be viewed as a small trainable peak surround suppression algorithm.

Finally, in a very similar paper [20], authors use the sequence of Hough Transform, several convolutions and an inverse Hough Transforms as a part of their network for pixel-based segmentation of straight line segments. From the perspective of a general-purpose line detection, the dataset used to train and test the network - Wireframe [16] is inconsistent, as some clearly salient straight lines are absent from the desired answer (see images below). We believe that the CNN proposed in [20] spends many parameters to distinguish between "important" and "unimportant" lines and, possibly, overfits to the dataset strongly.

The objective of this paper therefore is to propose a very lightweight CNN that can be seen as an enhancement of Hough line detector for general purpose.

The rest of the paper is organized as follows: in Section 2 we brief the reader on the basics of Hough transform and describe the general idea our approach. In Section 3, we discuss what disadvantages traditional convolutional neural networks (CNNs) have when applied to line detection; and how inserting a Hough transform solves these problems. In Section 4 we present the experimental setup: the exact architectures and the procedure for synthetic dataset generation used to train the networks. Section 5 provides the results and discussion.

## 2. LIMITATIONS OF TRADITIONAL CNNs FOR LINE DETECTION

Let us consider a traditional CNN that solves the task of line detection. For simplicity, we can assume that each output neuron of this CNN provides an answer whether some straight line in the image is "salient" or not. The receptive field of such a neuron (a set of input image pixels that affect the neuron) should cover the corresponding line and the pixels in its proximity. At the same time, in a traditional CNN built with convolutions and max pooling operators, all neurons have square receptive fields of the same size. This means that output neurons of a LD network will have square receptive fields that cover the whole image. This consideration lets us formulate two main weaknesses of traditional CNNs for line detection.

Global receptive field is necessary for efficient LD, but traditional CNN components are inefficient at obtaining it.
To get a global receptive field, we either have to stack many convolutions (which is not efficient) or use multiple max poolings (which degrades the resolution of the image and, subsequently, the accuracy of found line parameters). Naturally, there are some work-arounds to the mentioned problem. For example, a large receptive field can be achieved without resolution degradation by atrous convolutions [16] — reported to have poor performance for large dilation rates [21]. Alternatively, output resolution reduced by max pooling operators can be restored by upsampling [22]. Combination of pooling-upsampling CNNs have become a standard for LSD problem [16,22]. Unfortunately, such approaches are computationally expensive: e.g. in [22] a traditional CNN for line segment detection yields 10 FPS on a GPU (although this complexity cannot be solely attributed to the max pooling - upsample combination).

Output neurons of a traditional CNNs can only have a square/rectangular receptive field, which contains excessive context for the line detection problem.
The necessary context for the decision on whether some line on the image is salient is usually concentrated around that line. Ideally, a neuron that corresponds to this line should have a receptive field that covers only this strip-like area of interest — such a configuration would allow to reduce the number of parameters and to exclude the unnecessary context from computation. But for this to be the case, different output neurons should have receptive fields of different shapes and orientations.

And with a traditional CNN that is not possible: all neurons can only have square (or rectangular) receptive fields (with sides of the rectangular parallel to image sides). This is not an unnatural limitation for the mainstream "bounding box" detection problems (e.g. detection of persons or vehicles [23]), and therefore this particular weakness of CNNs has not been studied in detail. However, this is a significant limitation for the LD problem, where line can have any orientation.

By incorporating the Hough layer into the CNN we combat both outlined problems: we provide strip-like receptive fields spanning the whole image; additionally, after Hough layer, different neurons have receptive fields of different shapes and orientations. More information on receptive fields of the proposed networks is provided in Section 4.

## 3. HOUGH TRANSFORM

### 3.1 Overview of Hough Transform

HT is a classical instrument of computer vision originally aimed to detect lines [9, 10] and later generalized for the detection of patterns of arbitrary shapes [15]. It is usually applied for the detection of patterns that can be described by a small set of parameters. Hough transform-based detection operates by first accumulating the evidence for possible parameter values in the so-called accumulator space, and then finding peaks in it. In this paper, however, we view Hough transform simply as a function that takes an image and outputs sums of pixels intensities along each its straight line.

There are several established variations of Hough transform. These variations not only use different order of calculations, but also use different sets of pixels to approximate the lines in the image, which means they have different complexities and optimal inference algorithms.

For example, the brute force summation over all lines in an image has $O(N^3)$ complexity (where $N$ is the linear size of the image). In this paper we propose to use Fast Hough Transform (FHT) with $O(N^2 log N)$ complexity. FHT approximates the lines with dyadic patterns; and utilizes an efficient scheme of summation. The speedup is achieved by reusing the sums over the sub-patterns, which are included into multiple dyadic lines. More detailed information can be found in [15].

### 3.2. Our implementation of Fast Hough Transform

Our FHT implementation works on each channel of the input image (tensor) independently, so let us for simplicity assume that input image has 1 channel. Input image must be an *NxN* square, and *N* must be a power of two: $N = 2^n$. FHT processes this image and outputs 4 images, each with height *N* and width *2N-1*. Each pixel of these output images contains a sum over some line from the input image.

The output (sums over lines) is structured into four different channels based on the original inclination of the line relative to the Y-axis of the input image. Lines with following orientations are grouped together: *(-90, -45), (-45, 0), (0, 45), (45, 90)*. The output is structured this way as a necessary step of computational optimization.

The output tensor has width *2N-1* equal to the number of lines within the according orientation range. Intuitively, it may seem that there should be *N* lines within each category (one line per pixel on the side of the image), however, some of the lines can start "out of the image". Let us illustrate how these images are formed for lines with inclination in *(0,45)* range - refer to Fig. 3.

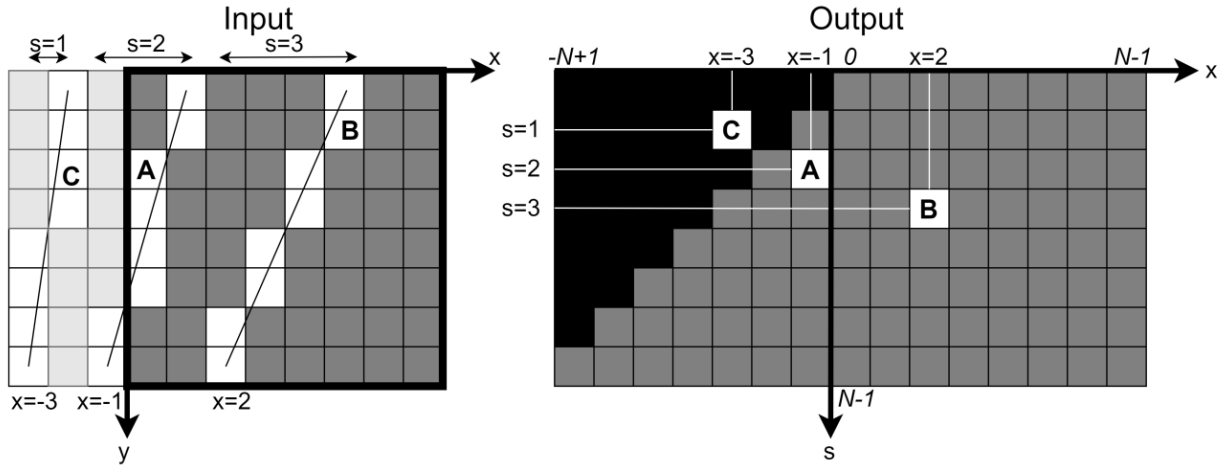

Figure 3. Hough transform (HT) illustration for an 8x8 image. Pixel *A(-1, 2)* in Hough space contains a sum of image intensities along the straight line *A* with offset *x=-1* and shift *s=2*, shown in gray.

The line is parametrized by coordinate *x* of the "bottom point" of the line (which may fall out of the image) and by the shift *s* - the absolute difference between *x* coordinates of top and bottom points of the line. Since one image contains lines within 45 degrees range, *s* takes values from *0* to *N*. Coordinate *x* can take values from *-(N-1)* to *(N-1)* -- hence the *(N x 2N-1)* shape of the output.

In the "Input" part of the Fig. 3, line *A* has *x=-1* and *s=2*; line *B* has *x=2* and *s=3*. Pixels that store sums over these lines are shown in the "Output" part of Fig. 3.

Now, let us consider line *C (x=-3, s=2)*. As seen from the input image, the line lies completely out of the image. The sum for such line should always be 0. There are a total of $N^2/4$ such pixels in each output image; specifically, all lines with *x<0* and *|x| > s* lie completely out of the image and have zero sum. In the output part of the figure, the area that stores such pixels is colored in black.

This variation of HT (as any other) consists of only summations over sets of image pixels and therefore is fully differentiable and can be inserted into a neural network without the loss of end-to-end trainability.

### 3.3. Rationale behind convolutions in Hough space

An idea to embed HT into a CNN was first proposed in [17] for the problem of vanishing points detection. The proposed architecture was *conv - HT - conv - HT - conv,* where *conv* denotes any block of several convolutional layers.

With such an architecture, some of the convolutions are conducted in Hough space (instead of in the image *(x,y)* space). Let us consider why convolutions are effective for the analysis of Hough space images.

There are two main reasons why convolutions are widely used for image analysis: firstly, neighboring pixels in an image are strongly correlated, while far pixels are weakly correlated. Using convolutions naturally limits the distance across which the pixel dependencies are modeled. Secondly, with convolutions we reuse the same weights at every position of the image - it is parameter-efficient and guarantees prediction invariance to the shifts of objects within the input image.

Convolutions hold similar advantages after transition into Hough space. Firstly, neighboring points in Hough space correspond to very similar lines in an image (almost parallel and/or close to each other) -- therefore their responses are correlated (while far points correspond to lines that strongly differ by their positions or orientations).

Secondly, consider that the goal of the Hough layer is to find "salient" lines in the image; and whether the line is salient should not depend on its rotation and position. This requirement is automatically met when convolutions in Hough space are used: indeed, translation of the input image results in horizontal translation of its Hough image; rotation of the input image results in vertical translation of its Hough image (although rotation also causes irregular stretching of Y axis of Hough image, it is slight and may be neglected).

## 4. PROPOSED APPROACH

With the proposed approach, output of our network is an image, with each pixel corresponding to some line in the input image. The overview of the proposed approach is presented in Fig. 2. The first block of convolutions, *convA*, operates in the image space. Then HT aggregates *convA* output along straight lines, providing strip-like receptive fields for the next block of convolutions, *ConvB*. *ConvB* is followed by a non-maximum suppression (NMS).

LNet can be viewed as a trainable variation of a classical approach to the line detection [12,15]: (i) extract feature map from the image, (ii) apply Hough transform over the feature map, (iii) locate and refine peaks in Hough space. In the non-CNN approaches, the first step was traditionally performed by the Canny edge detector and the third step by engineered filters and NMS. In LNet, we substitute them with convolutional blocks: convA and convB (with NMS), correspondingly.

### 4.1 CNN architecture

We report two very lightweight CNNs: LNetFast and - slower but more accurate - LNetAcc. They differ in the number of convolutions and convolution channels in blocks *convA* and *convB*. Compared to some of the CNNs used in literature for LD and LSD, both are extremely light - both consist of a small (3 and 6 convolutions, respectively) number of layers with small (1 to 8) number of channels and a FHT. The detailed architectures of the networks, the numbers of parameters and computational complexities for each layer are provided in Tables 1 and 2.

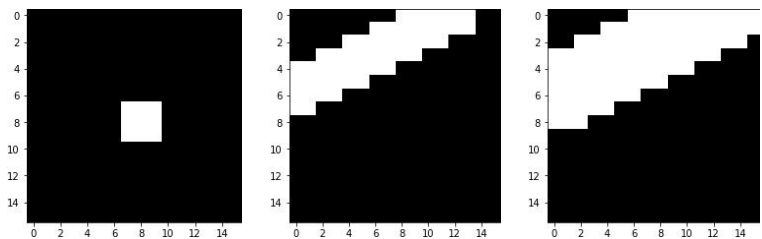

Figure 3. LNet's receptive fields for a 16x16 image. Left – convA, middle – HT, right – convB (output).

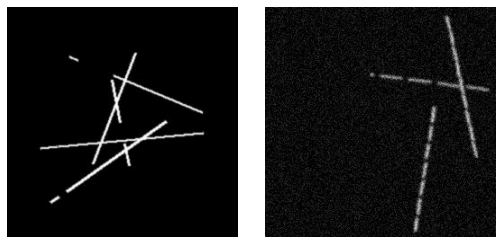

Figure 4. Samples from the synthetic dataset used.

| block | Convolutional layers | | | # param | MFLOP for 256x256 input |
|---|---|---|---|---|---|
| | kernel, NxHxWxC | pad | dilation | | |
| convA | 1x3x3x1 | 1 | 1 | 10 | 1.2 |
| HT | | | | | 4.1 |
| convB | 4x3x3x1 | 1 | 1 | 40 | 19.9 |
| | 1x1x1x4 | 0 | 1 | 5 | 2.4 |
| Total | 3 convolutional layers | | | 55 | 27.6 |

Table 1. LNetFast architecture. All convolutions are followed by ReLU. MFLOP accounts for all branches. Addition and multiplication are treated as equal FLOP.

| block | Convolutional layers | | | # param | MFLOP for 256x256 input |
|---|---|---|---|---|---|
| | kernel, NxHxWxC | pad | dilation | | |
| convA | 4x3x3x1 | 1 | 1 | 40 | 5.0 |
| | 1x3x3x4 | 1 | 1 | 37 | 4.8 |
| HT | | | | | 4.1 |
| convB | 8x3x3x1 | 1 | 1 | 80 | 39.8 |
| | 8x3x3x8 | 2 | 2 | 584 | 304.1 |
| | 8x3x3x8 | 3 | 3 | 584 | 304.1 |
| | 1x1x1x8 | 0 | 1 | 9 | 4.5 |
| Total | 6 convolutional layers | | | 1334 | 666.4 |

Table 2. LNetAcc architecture. All convolutions are followed by ReLU. MFLOP accounts for all branches. Addition and multiplication are treated as equal FLOP.

### 4.2. Receptive field

Let us consider LNet's receptive field. Here by "receptive field of a neuron" we mean "a set of input image pixels, such that their values affect the value of the neuron", by "receptive field of a layer" (or that of the whole network) we mean a receptive field of a layer's neuron. The receptive fields of different LNet's layers are illustrated in Fig 3.

The receptive field of *convA* is a 3x3 square – a result of one convolutional layer. Hough transform aggregates *convA* output along straight lines so one neuron in Hough space has a straight strip-like receptive field aligned with the line associated with that neuron. Convolutions in the Hough space aggregate information along lines that are close and have similar orientation - so the receptive field of *convB* is additionally stretched and widened.

The proposed LNet architecture solves both problems of traditional CNNs for line detection, outlined in Section 2. Firstly, LNets provide global (spanning the whole image) receptive field, without the excessive amount of maxpool stacking; secondly, each output neuron has different receptive field – depending only on the pixels near the line that it is designed to detect and does not include the unnecessary context filling the rest of the input image. LNet provides natural prediction encoding: for each straight line in the image, there is exactly one corresponding output neuron.

## 5. DATASET AND EXPERIMENTAL SETUP

### 5.1. Dataset

Convolutional neural networks have been applied to the line detection problem only recently; so it is natural there are only few existing datasets. There is Wireframe dataset [16], however the markup for some images is inconsistent (some papers citing it aim at the "artistically important line detection") – so it cannot be used as a general line-detection algorithm. There is also a document detection [24] dataset and York [25] dataset, both sharing the same problem of markup.

Figure 5 shows two examples for WireFrame dataset (on the left). See how the reflections of the pillars are not marked up, although they look the same as original columns; the reflection of the long black line spanning the whole border of the pool is also not marked up. On the next image, the drawing on the wall consists of numerous very salient straight segments, which are not marked-up; the very contrastive line between the bed sheets is also not marked up, while the barely visible line on the top curtain is marked-up.

On the right, the image from York dataset and its markup is shown. All lines of bricks are exactly the same, however, only some of them are marked up. The stone on the ground is not marked up either. Such problems are present in most images of the datasets, especially in the York one. It is possible that the CNNs that achieve top quality on these dataset "spend" some of their representational power to overfit to these inconsistencies in datasets.

Since our goal was to train a lightweight "Hough enhancer" network, we decided to train and test it on a simple synthetic dataset (Fig. 4), where the ground truth is unambiguous and well-defined. The generated dataset contains a total of 1000 256x256 grayscale images (800 train and 200 test). Each image was generated randomly and independently. Images vary in number of lines, their locations and appearance (solid, dotted, etc.), additive Gaussian noise and line blurring.

Routine for generating images:
1. Take empty (black) image.
2. Add a *uniform(1,5)* number of segments into the image. The lines containing these segments constitute the ground truth for this image.
3. Each segment is with equal probability dense, dotted or complex.
    3.1. Dense segment is simply a segment filled with white pixels.
    3.2. Dotted segment consists of periodic subsegments with random period *T* distributed as *uniform(0.07, 0.25)* of the total segment length. Within the period, the relative length of white part is *uniform(0.6, 0.9)*.
    3.3. Complex segment consists of *uniform(2,5)* number of random subsegments. They may overlap, but the minimum of 2 non-overlapping subsegments is required (they are randomly resampled until this condition is satisfied).
6. Add independent uniform intensity noise. The amplitude of this noise is determined per-image and distributed as *uniform (0, 0.25)*.
7. Use Gaussian blur on the image with *uniform(0,1.5)* kernel size.

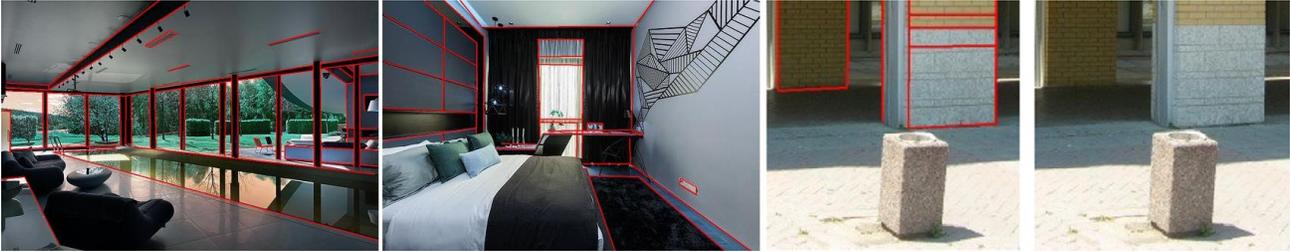
Figure 5. Major inconsistencies in existing line detection datasets (Wireframe –two left images, York – two right images).

## 5.2. Training
Weight initialization

To speed-up training and avoid bad minima, we initialize LNet weights to provide a reasonable starting quality. We assume that it is reasonable to initialize the weights so that the network performs exactly HT (up to a linear transform). This was done by initializing all convolutional kernels with an identity transformation (1 in the center and 0 elsewhere).

Unfortunately, such an initialization is not suitable: with it, all the kernels of each layer maintain the exact same weights (even during training), which results in a degenerate expressive power of the network.

To deal with this issue, we decouple the kernels' parameters by adding noise to the initial HT-like initialization. Noise is generated with Kaiming uniform distribution [26] multiplied by $10^{-2}$. Our experiments have shown such initialization to be the most stable (the networks perform better on average and do not degrade into "all zeros").

Loss function and optimizer

Let us consider a ground truth image: an image in Hough space with zeros everywhere but the points that correspond to the sought-for lines. We used Gaussian blur (with $\sigma = 1.8$) to acquire *target* image that was used for training.

We used pixel-wise weighted Mean Squared Error (MSE) between LNet output and *target* as the loss function. The weighting is necessary because most points in Hough space are zeros and with regular MSE the network provided trivial answers ("there are no lines"). To stimulate low error around actual lines, we weighted (pixels-wise) MSE with *1+1000*target*.

We trained LNet for 30 epochs with Adam optimizer with learning rate $lr = 10^{-3}$, halved each 10 epochs. Batch size was 32, $L_2$ weight decay set to $10^{-5}$.

## 5.3. Quality Evaluation
To evaluate the detection quality, we use the *distance* between two lines, similar to the distance between line segments [16]: (1) find ends as the points of intersection between each line and image frame, (2) match the ends of one line to those of the other, (3) compute the distance between lines as the mean distance between matching ends.

To calculate the quality, we:
1. Drop the detected lines with confidences below a *confidence threshold*.

2. For each ground truth line, check if there is a detected line within distance threshold. If yes, we consider this a *true positive* and remove both lines from the procedure. If there are multiple detected line within the threshold, line with maximum confidence is chosen.
3. The unmatched detected lines are considered *false positive*; the unmatched ground truth line are considered *false negative*.

To measure the quality we first calculate traditional precision-recall curve, and then derive scalar metrics: precision@90recall, recall@90precision, AP (average precision).

### 5.4. Results

As a baseline detector we use Hough Transform followed by NMS. We tried alternative baselines: standard HoughLines algorithm from OpenCV library [27], with various preprocessing methods: Otsu binarization [28], Canny edge detection [29], skeletonization [30]. They did not provide a quality gain over the straightforward Hough Transform plus NMS, so we used Hough as the main baseline.

Comparison of LNetFast and LNetAcc with the baseline is presented in Table 3. The reported LNets' qualities are averaged (median) over 5 runs. Both LNetFast and LNetAcc outperform the baseline by a large margin. The rise in MFLOP could appear significant, but it assumes sequential evaluation, while convolutions (as well as Hough Transform itself) can be vectorized efficiently on GPU. Therefore, we expect the actual speed of LNets' to be suitable for real-life applications.

| method | AP | precision@90recall | recall@90precision | MFLOP for 256x256 input |
|---|---|---|---|---|
| baseline | 91.63 | 91.27 | 93.85 | **4.1** |
| LNetFast | 94.74 | 93.13 | 97.23 | 27.6 |
| LNetAcc | **95.49** | **93.60** | **98.61** | 666.4 |

*Table 3. Metrics of two LNets and Hough baseline.*

### 5.5. Discussion

In this paper we have considered a very simplistic dataset. Our goal was to illustrate the advantage of our method over the classical HT. We expect the proposed approach applied to a real-world dataset to preserve its key properties. We avoided using Wireframe or York datasets for training due to the mentioned inconsistencies in the markup.

Overall, one of the challenges of training a line detector is determining the ground truth: how long should the line be to be considered salient? How contrasted should it be? Does a straight border between two textured surfaces count as a line? The ambiguity of ground truth for line detection tasks is the reason we believe generating realistic synthetic datasets can benefit the development of trained line detectors. Our future work includes generation of feature maps more similar in distribution to the feature images obtained from real data.

### 6. CONCLUSION

In this paper we propose a convolutional neural network with Hough layer for line detection. The algorithm is end-to-end trainable. We argue that the conventional CNNs have two significant disadvantages when applied to the task of line detection, related to the inflexibility of the receptive field: (i) traditional CNNs are inefficient at obtaining the receptive field spanning the whole image; (ii) all output neurons of traditional neural networks have the same square receptive field – which provides excessive amount of context and unnecessarily increases the amount of parameters. We show how inserting the Hough layer into the network provides the strip-like receptive fields and allows to solve these two problems.

We construct the synthetic dataset for line detection training, mainly because existing datasets – Wireframe and York have extremely inconsistent markup. We train the proposed networks and show their advantage over classical Hough transform-based detection.

## 7. ACKNOWLEDGMENTS

The research was partially supported by Russian Foundation For Basics Research (projects 18-29-26033, 18-29-26028).